# AraGenre 2026: A Hierarchical Definition-Guided Arabic Genre Classification Shared Task

**Mo El-Haj**[1,2], **Saad Ezzini**[3], **Shadi Abudalfa**[4,5],
**Mustafa Jarrar**[6], **Nguyen Minh Chi**[1] and **Nguyen Minh Quan**[1]

[1]VinUniversity, Vietnam; [2]Lancaster University, UK
[3]King Fahd University of Petroleum and Minerals, Saudi Arabia
[4]Onaizah Colleges, Saudi Arabia, [5]University College of Applied Sciences, Palestine
[6]Hamad Bin Khalifa University, Qatar
m.el-haj@lancaster.ac.uk

## Abstract

AraGenre is a shared task on hierarchical, definition-guided Arabic genre classification, motivated by the limited availability of annotated data in Arabic and other low-resource languages. Systems assign each Arabic text segment both a broad communicative genre and a fine-grained specific genre. The released training and development sets contain limited, primarily synthetic and controlled examples, whereas the hidden final benchmark contains noisier naturally occurring text spanning Modern Standard Arabic, Classical Arabic, and multiple dialects. Participants received natural-language definitions for 74 previously unseen specific genres, creating a zero-shot label generalisation setting in which systems had to infer class semantics rather than memorise fixed label–feature associations. The task attracted 46 registrations and 373 submissions, with 17 teams completing the final evaluation. Thaka ranked first with a Hierarchical Macro F1 of 0.7352, followed by HoangPhong (HP) with 0.7169 and NAMAA with 0.7013. The results show strong broad-genre recognition but a substantial gap in fine-grained classification under linguistic and domain variation.

## 1 Introduction

Genre captures a text's communicative purpose, discourse structure, and stylistic behaviour rather than only its subject matter. Genre classification supports information retrieval, corpus construction, content recommendation, educational technology, and downstream NLP. For Arabic, the task is complicated by diglossia, dialect variation, optional diacritisation, non-standard orthography, and the coexistence of formal, informal, historical, and contemporary writing. These challenges are particularly important in low-resource NLP, where annotated datasets are often small, domain-specific, and uneven across linguistic varieties. The importance of reusable language resources and evaluation frameworks has long been recognised (El-Haj et al., 2015). Recent work has expanded resources for Hindi (Lal et al., 2025), Igbo (Chukwuneke et al., 2023), Vietnamese (Huy et al., 2026; Dinh et al., 2026), and Welsh (Knight et al., 2024; Khallaf et al., 2023; Ezeani et al., 2022). However, resource availability does not guarantee generalisation beyond the classes, domains, and linguistic patterns observed during training.

AraGenre formulates Arabic genre classification as a hierarchical semantic generalisation task rather than a conventional closed-set problem. Systems receive natural-language genre definitions and must classify texts within an expanded label space containing previously unseen specific genres. The final evaluation also introduces naturally occurring text and substantial variation in domain, dialect, register, and orthography. Systems must therefore transfer knowledge from class descriptions and limited demonstrations rather than rely on memorised label–feature associations.

The shared task contributes: (1) a two-level Arabic genre taxonomy; (2) a definition-guided framework for zero-shot label generalisation; (3) a controlled low-resource training setup paired with a naturally occurring, distribution-shifted evaluation benchmark; (4) two reproducible AraGenre baselines; and (5) an official evaluation involving 17 final teams.

## 2 Motivation and Related Work

Genre-specific resources remain comparatively limited and often use small, fixed label inventories. El-Halees (2017); Elnagar et al. (2020) classified Arabic documents using stylistic and linguistic fea-tures, while Ferhat et al. (2024) introduced the Arabic Functional Text Dimensions Corpus, containing 3,400 documents across 17 categories. These resources provide useful supervised benchmarks but generally assume that all genres are known and represented during training. This assumption is difficult to maintain in low-resource and multilingual settings, where annotated data may be unavailable for particular dialects, domains, or emerging communicative practices. Work on Hindi (Lal et al., 2025), Igbo (Chukwuneke et al., 2023), Vietnamese (Huy et al., 2026; Dinh et al., 2026), and Welsh (Knight et al., 2024; Khallaf et al., 2023; Ezeani et al., 2022) highlights the continuing need for language-specific resources and models that remain robust beyond observed training distributions. AraGenre addresses this challenge by incorporating label semantics into the model input and expanding the fine-grained genre inventory during evaluation. The task is related to label-description learning (Gao et al., 2023) and zero-shot hierarchical classification (Paletto et al., 2024). Unlike fixed-label benchmarks, AraGenre evaluates transfer to 74 previously unseen specific genres represented only by their names, definitions, and positions within the broad taxonomy. The benchmark therefore tests whether systems can model communicative function rather than rely on lexical memorisation. Success requires aligning Arabic text representations with English genre definitions, exploiting hierarchical relationships, and remaining robust under distribution shift.

## 3 Data Collection and Creation

Each instance consists of an Arabic text segment ranging from a short fragment to a longer passage. The released training and development sets contain primarily synthetic and carefully controlled examples designed to provide limited supervision. In contrast, the hidden final benchmark contains substantially noisier, naturally occurring texts drawn from diverse communicative settings. The data cover Modern Standard Arabic, Classical Arabic, and multiple Arabic dialects, including formal and informal writing as well as diacritised and undiacritised text.

The source texts were collected from corpora available through the ArabicNLP.uk repository.[1] These include Tarab (El-Haj, 2026), AraFinNews (El-Haj and Rayson, 2025), ArabJobs (El-Haj, 2025), the Multilingual Corpus of World's Constitutions (MCWC) (El-Haj and Ezzini, 2024), KALIMAT (El-Haj and Koulali, 2013), HABIBI (El-Haj, 2020), the Essex Arabic Summaries Corpus (EASC) (El-Haj et al., 2010), the Arabic Dialects dataset (El-Haj et al., 2018), DARES (El-Haj et al., 2024), and the MultiLing corpora (Li et al., 2013). Together, these resources provide broad coverage of domains and communicative settings, including news, finance, recruitment, literature, song lyrics, online discussion, dialectal writing, and summarisation data.

The hierarchy contains six broad genres: *Informative*, *Creative*, *Interactive*, *Learning*, *Legal*, and *Religious*. Each broad category contains a set of finer-grained communicative genres.

Appendix B's Table 3 presents representative genre labels and shortened definitions, while Table 4 provides further examples on specific genres. Together, they show that the taxonomy covers distinct communicative functions, ranging from analytical and educational writing to motivational, diplomatic, and religious discourse, while also acknowledging potential overlap with related linguistic concepts such as register, text type, and discourse function.

Natural-language definitions were created for both the broad and specific genres. Initial drafts were generated using ChatGPT 5 (Singh et al., 2025) and subsequently reviewed, corrected, and manually post-edited by the task organisers. This process ensured that each definition accurately represented the intended communicative function, distinguished the genre from neighbouring categories, and remained sufficiently informative for definition-guided classification.

The final evaluation introduced 74 specific genres that had not appeared as labelled classes in the released data. Participants received their English names and definitions, but no labelled examples for these genres. For example, *Sudanese song lyrics* was defined as expressive and rhythmic Sudanese Arabic intended for musical performance, while *administrative and secretarial jobs* covered recruitment texts describing office duties, scheduling, documentation, communication skills, and workplace requirements. The split therefore tests transfer across genres, domains, dialects, and writing conventions (See table 4 in Appendix B).

[1] https://arabicnlp.uk/

## 4 Task Description and Evaluation

For each input text, a system must output both a broad genre label and a specific genre label:

$$(\hat{y}^{\text{broad}}, \hat{y}^{\text{specific}}). \tag{1}$$

Each specific genre belongs to one of the six broad genre branches. Nevertheless, broad and specific predictions are scored independently, meaning that the scorer does not enforce hierarchical consistency between the two predicted labels.

The shared task was deliberately designed to contrast limited development supervision with a substantially larger and more diverse final evaluation. As summarised in Table 1, participants initially received only 140 labelled training examples, covering seven specific genres from five broad genres. The development benchmark contained 110 examples covering six specific genres and all six broad genres. This restricted setup simulated a low-resource scenario in which systems had access to few labelled examples and only a narrow portion of the complete genre space.

| Split | Instances | Broad | Specific |
|---|---|---|---|
| Training | 140 | 5 | 7 |
| Development | 110 | 6 | 6 |
| Final evaluation | 27,972 | 6 | 74 |

Table 1: Number of instances and distinct genre labels in each AraGenre split.

The final evaluation introduced a considerably more challenging setting. It contained 27,972 naturally occurring Arabic text segments and 74 previously unseen specific genres. Participants received the genre names and natural-language definitions but no labelled examples for these categories. The task therefore required zero-shot label generalisation from a small development setting to a substantially expanded label inventory and a large out-of-distribution evaluation set.

The 74 specific genres were distributed unevenly across the hierarchy: 42 belonged to *Informative*, 12 to *Learning*, seven to *Creative*, six to *Religious*, four to *Interactive*, and three to *Legal*. The final inventory covered diverse communicative settings, including poetry and dialectal song lyrics, news, job advertisements, book descriptions, encyclopaedic writing, online reviews and comments, student writing, textbooks, legal documents, and religious texts. Examples include *Sudanese song lyrics*, *administrative and secretarial jobs*, *Twitter posts*, *early education textbooks*, *constitutions*, and *tafsir*.

This design evaluates whether systems can transfer from limited demonstrations to unseen categories by aligning the semantic content of Arabic texts with English genre definitions. It also tests robustness to changes in genre inventory, domain, dialect, register, orthography, and dataset scale. Development and final scores should therefore not be interpreted as estimates drawn from the same data distribution.

The official scorer matches predictions to the gold data using the instance identifier. All gold identifiers are evaluated, including those for which a participant provides no prediction. Missing instances or missing broad or specific labels are assigned the special value INVALID and are consequently penalised. Predictions containing identifiers that do not occur in the gold data are ignored. When a submission contains duplicate identifiers, only the final prediction associated with each duplicated identifier is retained. This prevents systems from submitting multiple candidate labels for the same instance.

Performance is calculated independently for broad and specific genres using Macro F1, Weighted F1, and accuracy. Macro F1 assigns equal importance to each evaluated genre, regardless of its frequency, while Weighted F1 weights each genre according to its number of gold instances. The implementation uses scikit-learn with undefined class-wise F1 values set to zero. The official ranking metric is Hierarchical Macro F1:

$$F_{\text{hier}} = \frac{F_{\text{macro}}^{\text{broad}} + F_{\text{macro}}^{\text{specific}}}{2} \tag{2}$$

The metric gives equal weight to high-level communicative classification and fine-grained genre recognition. This is particularly important because a system may correctly identify the broad communicative function while failing to distinguish between closely related specific genres. Predictions were submitted in the official JSON format and evaluated automatically through Codabench.[2]

## 5 Baselines

Two AraGenre baselines were released as reproducible reference systems. **Baseline 1** embeds each input text and genre definition using a multilingual sentence encoder and selects labels according to semantic similarity, following the sentence-embedding paradigm of Reimers and Gurevych (2019). **Baseline 2** uses retrieval-augmented generation with an instruction-tuned multilingual language model, combining retrieved demonstrations and genre definitions within the prompt (Lewis et al., 2020). The starter kit additionally provides data loaders, evaluation scripts, example submissions, and Codabench-compatible prediction formatting.

[2] https://www.codabench.org/competitions/16356/

| R | Team | Hier. | Spec. | Broad |
|---|---|---|---|---|
| 1 | Thaka (Alqaeri et al., 2026) | **.7352** | **.5725** | .8979 |
| 2 | HP (Phong and Thin, 2026) | .7169 | .5339 | **.9000** |
| 3 | NAMAA-1 (Barmandah et al., 2026) | .7013 | .5114 | .8913 |
| 4 | GHADAI (Alsubhi and Kurdi, 2026) | .6050 | .4297 | .7803 |
| 5 | Nasaq-1 (Altamimi et al., 2026) | .6006 | .4264 | .7748 |
| 6 | phoenix (Chi, 2026) | .5800 | .4109 | .7492 |
| 7 | SAGE (Mansour et al., 2026) | .5278 | .3505 | .7051 |
| 8 | Mohannad Hendi (MohannadHendi, 2026) | .4560 | .2995 | .6125 |
| 9 | WathiqTeam (Bostaty et al., 2026) | .4559 | .2869 | .6250 |
| 10 | rsonbol | .4421 | .2724 | .6119 |
| 11 | Rabee Al-Qasem (Al-Qasem, 2026) | .4399 | .2511 | .6288 |
| 12 | Nasaq-2 (Altamimi et al., 2026) | .3580 | .1714 | .5447 |
| 13 | NAMAA-2 (Barmandah et al., 2026) | .3555 | .2129 | .4981 |
| 14 | NasSem | .3367 | .1930 | .4804 |
| 15 | Northwestern (Shlkamy and Zaghouani, 2026) | .3302 | .2008 | .4596 |
| 16 | GenreMinds | .3286 | .1923 | .4650 |
| 17 | morooj_6 | .3152 | .1894 | .4410 |
| – | AraGenre baseline | .2472 | .1358 | .3585 |

Table 2: Official final results. Hier., Spec., and Broad denote Hierarchical, Specific, and Broad Macro F1.

On the development benchmark, Baseline 1 achieved a Hierarchical Macro F1 of 0.4658, compared with 0.3899 for Baseline 2. Baseline 1 was also evaluated in the final phase, where it obtained 0.2472. This reduction should be interpreted in light of the deliberate change in evaluation conditions: the final benchmark increased from six development genres to 74 previously unseen specific genres, expanded from 110 to 27,972 instances, and replaced controlled examples with substantially noisier naturally occurring Arabic text.

## 6 Participation and Results

AraGenre received 46 registrations and 373 submissions. Seventeen teams submitted final predictions. The exported leaderboard did not include mandatory system fact sheets, so this overview reports verified official scores without assigning unconfirmed architectures or external resources to individual teams.

Thaka ranked first because its stronger specific-genre score outweighed HP's slightly higher broad-genre score. All 17 teams exceeded the organiser baseline. Across participating teams, mean Broad Macro F1 was 0.6509, whereas mean Specific Macro F1 was 0.3238, producing an average gap of 0.3271. Systems therefore identified the general communicative family much more reliably than the fine-grained genre.

The distribution shift was also substantial. For the 13 final teams whose usernames could be matched to development submissions, the mean best development score was 0.9344, while their mean final score was 0.5261, a reduction of 0.4082. The two phases are intentionally different and should not be treated as samples from one distribution. Nevertheless, the contrast demonstrates that near-ceiling performance on controlled examples does not guarantee generalisation to unseen genres and noisy naturally occurring text.

Particularly challenging cases are expected to include neighbouring definitions, short context-poor segments, dialect- or domain-specific lexical cues, and texts expressing more than one communicative function. Confirmed instance-level error analysis requires participant predictions and system descriptions and is therefore left to the corresponding system papers.

## 7 Conclusion

AraGenre introduced hierarchical, definition-guided Arabic genre classification under realistic linguistic and distributional variation. The task combined six broad genres with a changing fine-grained label space and tested final systems on 74 unseen specific genres represented only through English definitions. Seventeen teams completed the final evaluation, with Thaka achieving the highest Hierarchical Macro F1 of 0.7352.

The broad–specific performance gap and the development-to-final reduction show that fine-grained semantic generalisation remains challenging. Future editions should expand naturally occurring labelled data, provide parallel Arabic and English definitions, investigate hierarchy-aware consistency constraints, and support systematic analysis across dialects, text lengths, and overlapping communicative functions.

## Limitations

The development and final sets differ intentionally in naturalness and label coverage, so their scores are not directly comparable as estimates from a single distribution. The leaderboard also lacks standardised system descriptions, limiting controlled comparison of architectures, external resources, and computational cost, while genre boundaries remain interpretative and some texts may plausibly express more than one communicative function.

# A Participating Team System Descriptions

This appendix provides brief descriptions of the approaches developed by participating teams that submitted system description papers for the AraGenre 2026 Shared Task. These descriptions complement the official evaluation results reported in Table 2 and provide additional insight into the modelling strategies adopted for hierarchical Arabic genre classification.

## A.1 Thakaa

Thakaa uses a family-restricted, definition-guided LLM judge followed by agreement-gated correction, in which two instruction models re-predict labels while viewing the full 74-genre taxonomy, together with a within-family attractor drain that redistributes predictions toward under-predicted genres (Alqaeri et al., 2026).

### A.2 HP

The HP team uses Gemini 3.5 Flash Lite to pseudo-label the test set with definition-guided few-shot prompting, then fine-tunes MARBERTv2 and ARBERTv2 on approximately 28,000 generated labels for classification (Phong and Thin, 2026).

### A.3 NAMAA Community

NAMAA replaces poorly generalising fine-tuned encoders with zero-shot DeepSeek classification using official definitions, then improves robustness through three-sample self-consistency voting and disagreement resolution, including a Claude-assisted final pass for disagreements (Barmandah et al., 2026).

### A.4 GHADAI

GHADAI combines a fine-tuned multilingual E5 bi-encoder for definition-based retrieval with Qwen2.5-32B reranking of shortlisted genres, using cyclic option rotation to reduce positional bias during LLM candidate scoring (Alsubhi and Kurdi, 2026).

### A.5 Nasaq

Nasaq introduces SynAnchor, generating Arabic genre prototypes from English definitions using GPT-4o, retrieving candidates with BGE-M3, and selectively invoking GPT-5 to adjudicate ambiguous cases among the highest-ranked genres (Altamimi et al., 2026).

### A.6 Phoenix

Phoenix combines multi-view multilingual retrieval with transductive per-genre z-normalisation, then reranks shortlisted candidates using calibrated Qwen3-8B log-probabilities, score fusion, sibling reranking, metric-aware decoding, and margin-gated candidate expansion for final predictions (Chi, 2026).

### A.7 SAGE

SAGE fine-tunes sentence encoders for genre similarity and trains a residual adapter to refine definition embeddings, reducing the representational gap between English genre descriptions and corresponding Arabic text instances (Mansour et al., 2026).

### A.8 MohannadHendi

MohannadHendi fine-tunes multilingual E5-large contrastively on Arabic texts paired with five deterministic views of English genre definitions, then predicts labels by nearest-definition retrieval in a shared embedding space during inference (MohannadHendi, 2026).

### A.9 WathiqTeam

WathiqTeam combines multilingual E5, Arabic prototype retrieval, and BM25 within a broad-first architecture, using class-balanced pooling, reciprocal rank fusion, hierarchical gating, and selective Qwen3 reranking for uncertain predictions (Bostaty et al., 2026).

### A.10 Rabee Al-Qasem

Rabee Al-Qasem combines multilingual MPNet embeddings with genre-definition representations, a Conformer projection module, auxiliary broad-genre supervision, synthetic data augmentation, and seed-based ensembling for hierarchical Arabic genre classification and stable predictions (Al-Qasem, 2026).

### A.11 Northwestern

Northwestern formulates classification as zero-shot dense retrieval, encoding Arabic texts and English genre definitions with multilingual BGE-M3 and assigning labels through cosine similarity without task-specific fine-tuning or additional supervised training (Shlkamy and Zaghouani, 2026).

## B Examples of Broad and Specific Genre Labels

| Genre pair | Shortened definition |
|---|---|
| **Informative**<br>Analytical reports | Factual interpretation of trends, policies, or evidence. |
| **Interactive**<br>Advice columns | Seeking or providing practical, emotional, or social advice. |
| **Learning**<br>Educational explanations | Structured clarification of academic, scientific, or technical concepts. |
| **Legal**<br>Diplomatic communication | Official communication on international relations or cooperation. |
| **Creative**<br>Motivational writing | Expressive writing intended to inspire or encourage. |
| **Religious**<br>Religious commentary | Reflective explanation of religious, moral, or spiritual guidance. |

Table 3: Illustrative broad–specific genre pairs and shortened definitions.

| **Broad Genre** | **Specific Genre** | **Definition** | **Example 1** | **Example 2** |
|---|---|---|---|---|
| Informative | Analytical Reports | Structured analytical writing discussing trends, developments, policies, or evidence-based observations using factual interpretation or comparative analysis. | تشير البيانات الاقتصادية الأخيرة إلى ارتفاع معدلات التضخم مقارنة بالعام السابق، مما أدى إلى تراجع القوة الشرائية للأسر. | أظهرت الدراسة أن الاستخدام المفرط للهواتف الذكية بين المراهقين يرتبط بانخفاض ساعات النوم وضعف التركيز الأكاديمي. |
| Interactive | Advice Columns | Texts where individuals seek or provide practical, emotional, social, or moral advice regarding personal experiences or everyday situations. | أشعر بالتوتر الشديد قبل الامتحانات، ولا أعرف كيف أتعامل مع هذا القلق المستمر، فماذا تنصحونني؟ | أعاني من صعوبة في تنظيم وقتي بين العمل والدراسة، وأحتاج إلى نصائح تساعدني على تحقيق التوازن. |
| Learning | Educational Explanations | Explanatory educational texts intended to help learners understand academic, scientific, technical, or conceptual topics through clarification and structured reasoning. | يحدث التبخر عندما تتحول المادة من الحالة السائلة إلى الحالة الغازية نتيجة اكتسابها للطاقة الحرارية. | لحساب مساحة المثلث، نقوم بضرب طول القاعدة في الارتفاع ثم نقسم الناتج على اثنين. |
| Legal | Diplomatic Communication | Official communication concerning international relations, negotiations, cooperation, agreements, or political coordination between states or organisations. | أكدت الدول المشاركة أهمية تعزيز التعاون الإقليمي لمواجهة التحديات الاقتصادية المشتركة. | عقد الوفدان اجتماعاً ثنائياً لبحث آليات تطوير العلاقات التجارية بين البلدين خلال المرحلة المقبلة. |
| Creative | Motivational Writing | Expressive or reflective writing intended to inspire, encourage, or persuade readers through motivational rhetoric or self-improvement themes. | لا تسمح للفشل أن يوقفك، فكل تجربة صعبة تمنحك فرصة جديدة للنمو والتعلم. | النجاح لا يأتي صدفة، بل يبدأ بخطوة صغيرة وإيمان مستمر بقدراتك. |
| Religious | Religious Commentary | Reflective or explanatory religious writing discussing moral lessons, spiritual guidance, interpretation, or ethical values within a religious context. | يدعو الإسلام إلى الصبر والتسامح، ويحث الإنسان على معاملة الآخرين بالرحمة والعدل. | يوضح الكاتب أهمية الإخلاص في العمل، وأن النية الصادقة أساس قبول الأعمال عند الله. |

Table 4: Example broad and specific genres, their definitions, and illustrative Arabic samples.